# Danger Theory: The Link between AIS and IDS?

U Aickelin, P Bentley, S Cayzer, J Kim, J McLeod

University of Nottingham, uxa@cs.nott.ac.uk
University College London, P.Bentley@cs.ucl.ac.uk
HP Labs Bristol, Steve_Cayzer@hplb.hpl.hp.com
King's College London, Jungwon@dcs.kcl.ac.uk
University of the West of England, Julie.Mcleod@uwe.ac.uk

**Abstract** We present ideas about creating a next generation Intrusion Detection System (IDS) based on the latest immunological theories. The central challenge with computer security is determining the difference between normal and potentially harmful activity. For half a century, developers have protected their systems by coding rules that identify and block specific events. However, the nature of current and future threats in conjunction with ever larger IT systems urgently requires the development of automated and adaptive defensive tools. A promising solution is emerging in the form of Artificial Immune Systems (AIS): The Human Immune System (HIS) can detect and defend against harmful and previously unseen invaders, so can we not build a similar Intrusion Detection System (IDS) for our computers? Presumably, those systems would then have the same beneficial properties as HIS like error tolerance, adaptation and self-monitoring. Current AIS have been successful on test systems, but the algorithms rely on self-nonself discrimination, as stipulated in classical immunology. However, immunologist are increasingly finding fault with traditional self-nonself thinking and a new 'Danger Theory' (DT) is emerging. This new theory suggests that the immune system reacts to threats based on the correlation of various (danger) signals and it provides a method of 'grounding' the immune response, i.e. linking it directly to the attacker. Little is currently understood of the precise nature and correlation of these signals and the theory is a topic of hot debate. It is the aim of this research to investigate this correlation and to translate the DT into the realms of computer security, thereby creating AIS that are no longer limited by self-nonself discrimination. It should be noted that we do not intend to defend this controversial theory per se, although as a deliverable this project will add to the body of knowledge in this area. Rather we are interested in its merits for scaling up AIS applications by overcoming self-nonself discrimination problems.

## 1. Introduction

The key to the next generation Intrusion Detection System (IDS) ([9], [25], [26]) that we are planning to build is the combination of recent Artificial Immune System (AIS) / Danger Theory (DT) models ([1], [4], [5], [32]) with our growing understanding of cellular components involved with cell death ([3], [11], [31]). In particular, the difference between necrotic ('bad') and apoptotic ('good' or 'planned') cell death, with respect to Antigen Presenting Cells (APCs) activation, is important in our proposed IDS. In the Human Immune System (HIS) apoptosis has a suppressive effect and necrosis a stimulatory immunological effect, although they might not actually be as distinct as currently thought.

In the IDS context, we propose to use the correlation of these two effects as a basis of 'danger signals'. A variety of contextual clues may be essential for a meaningful danger signal, and immunological studies will provide a framework of ideas as to how 'danger' is assessed in the HIS. In the IDS context, the danger signals should show up after limited attack to minimise damage and therefore have to be quickly and automatically measurable. Once the danger signal has been transmitted, the AIS should react to those artificial antigens that are 'near' the emitter of the danger signal. This allows the AIS to pay special attention to dangerous components and would have the advantage of detecting rapidly spreading viruses or scanning intrusions fast and at an early stage preventing serious damage.

## 2. AIS and Intrusion Detection

Alongside intrusion prevention techniques such as encryption and firewalls, IDS are another significant method used to safeguard computer systems. The main goal of IDS is to detect unauthorised use, misuse and abuse of computer systems by both system insiders and external intruders. Most current IDS define suspicious signatures based on known intrusions and probes [25]. The obvious limit of this type of IDS is its failure of detecting previously unknown intrusions. In contrast, the HIS adaptively generates new immune cells so that it is able to detect previously unknown and rapidly evolving harmful antigens [28].

In order to provide viable IDS, AIS must build a set of detectors that accurately match antigens. In current AIS based IDS ([9], [12], [19], [13]), both network connections and detectors are modelled as strings. Detectors are randomly created and then undergo a maturation phase where they are presented with good, i.e. self, connections. If the detectors match any of these they are eliminated otherwise they become mature. These mature detectors start to monitor new connections during their lifetime. If these mature detectors match anything else, exceeding a certain threshold value, they become activated. This is then reported to a human operator who decides whether there is a true anomaly. If so, the detectors are promoted to memory detectors with an indefinite life span and minimum activation threshold (immunisation) [27].

An approach such as the above is known as negative selection as only those detectors (antibodies) that do not match live on [13]. However, this appealing approach shows scaling problems when it is applied to real network traffic [26]. As the systems to be protected grow larger and larger so does self and nonself. Hence, it becomes more and more problematic to find a set of detectors that provides adequate coverage, whilst being computationally efficient. It is inefficient, to map the entire self or nonself universe, particularly as they will be changing over time and only a minority of nonself is harmful, whilst some self might cause damage (e.g. internal attack). This situation is further aggravated by the fact that the labels self and nonself are often ambiguous and even with expert knowledge they are not always applied correctly [24].

## 2.1 The Danger Theory

We now examine the biological basis for the self-nonself metaphor, and the alternative DT hypothesis. The HIS is commonly thought to work at two levels: innate immunity including external barriers (skin, mucus), and the acquired or adaptive immune system [28]. As part of the latter level, B-Lymphocytes secrete specific antibodies that recognise and react to stimuli. It is this matching between antibodies and antigens that lies at the heart of the HIS and most AIS implementations.

The central tenet of the immune system is the ability to respond to foreign invaders or 'antigens' whilst not reacting to 'self' molecules. In order to undertake this role the immune system needs to be able to discern differences between foreign, and possibly pathogenic, invaders and non-foreign molecules. It is currently believed that this occurs through the utilisation of the Major Histocompatability Complex (MHC). This complex is unique to each individual and therefore provides a marker of 'self'. In addition, the cells within the immune system are matured by becoming tolerised to self-molecules. Together, through the MHC and tolerance, the HIS is able to recognise foreign invaders and send the requisite signals to the key effector cells involved with the immune response.

The DT debates this and argues that there must be discrimination happening that goes beyond the self-nonself distinction because the HIS only discriminates 'some self' from 'some nonself'. It could therefore be proposed that it is not the 'foreignness' of the invaders that is important for immune recognition, but the relative 'danger' of these invaders. This theory was first proposed in 1994 [29] to explain current anomalies in our understanding of how the immune system recognises foreign invaders. For instance, there is no immune reaction to foreign bacteria in the gut or to food. Conversely, some auto-reactive processes exist, e.g. against self-molecules expressed by stressed cells. Furthermore, the human body (self) changes over its lifetime. Therefore, why do defences against nonself learned early in life not become auto-reactive later?

The DT suggests that foreign invaders, which are dangerous, will induce the generation of cellular molecules (danger signals) by initiating cellular stress or cell death [30]. These molecules are recognised by APCs, critical cells in the initiation of an immune response, which become activated leading to protective immune interactions. Overall there are two classes of danger signal; those which are generated endogenously i.e. by the body itself, and exogenous signals which are derived from invading organisms e.g. bacteria [16]. Evidence is accruing as to the existence of myriad endogenous danger signals including cell receptors, intracellular molecules and cytokines. A commonality is their ability to activate APCs and thus drive an immune response.

We believe that the DT will provide a more suitable biological metaphor for IDS than the traditional self-nonself viewpoint, regardless whether the theory holds for the HIS, something that is currently hotly debated amongst immunologists ([37], [21], [39]). In particular, the DT provides a way of grounding the response, i.e. linking it directly to the attacker and it removes the necessity to map self or nonself [1]. In our model, self-nonself discrimination will still be useful but it is no longer essential. This is because nonself no longer causes a response. Instead, danger signals will trigger a reaction. Actually, the response is more complicated than this, since it is believed that the APCs integrate necrotic ('danger') and apoptotic ('safe') signals in order to regulate the immune response. We intend to examine this integrative activity experimentally, which should provide useful inspiration for IDS.

**2.2  The DT in the Context of the HIS**

One of the central themes of the DT is the generation of danger signals through cellular stress or cell death. Cell death can occur in two ways; necrosis and apoptosis, and although both terminate in cell death, the intracellular pathways of each process are very distinct. Necrosis involves the unregulated death of a cell following cell stress and results in total cell lysis and subsequent inflammation due to the cell debris. Apoptosis, on the other hand, is a very regulated form of cell death with defined intracellular pathways and regulators [23]. Physiologically, apoptosis is utilised by the body to maintain tissue homeostasis and is vital in regulating the immune response. Once apoptosis is initiated extracellular receptors on the cell signal to phagocytic cells, e.g. APCs to remove the dying cell from the system.

Apoptosis can be initiated in a number of ways including; cytokine deprivation, death receptors e.g. CD95 and UV irradiation each having unique intracellular signalling profiles [17]. Interestingly, recent work has suggested that apoptotic pathways may not be as distinct from necrosis as previously assumed [20] and indeed may be inter-related. In both cases phagocytosis of the dying cell occurs and studies suggests that the APCs receive signals from the dying cells that affects activation state of the APCs themselves [35]. These results are of particular interest since they support the concept of danger signals, with the APCs being a rheostat responding to 'input' signals from cells undergoing necrosis, tipping the immune balance towards a pro-inflammatory state, which is an 'output' signal.

Evidence to support the critical role of cell death signals in APC activation has shown that APCs, which have phagocytosed necrotic cells, generate pro-inflammatory cytokines e.g. interleukin (IL) −1, interferon (IFN) and necrotic cells have been found to activate APCs in a vital step towards an immune response ([35], [15]). Of particular interest is the finding that cells undergoing apoptosis, rather than being invisible to the APCs, may actually help regulate the APCs response to necrotic cell debris. Studies [14] have shown that apoptotic cells actively down-regulate the APC activity by generating anti-inflammatory cytokines e.g. TGF and PGE2, although in other cases this has not been observed [35].

In a further complexity to the balance, reports have shown that necrotic and apoptotic cells work together to affect the APC activation and subsequent immune response [35]. Therefore a balance between cell death, either necrotic or apoptotic, would appear to be critical to the final immunological outcome. Here, we seek to understand how the APCs react to the balance of 'input' signals from apoptotic and necrotic cell death with an aim to determine and simplify the danger signal 'output'.

Previous studies have observed alterations in the generation of pro- and anti-inflammatory cytokines e.g. IL-1, IFN, TGF-ß and PGE2 following APC incubation with necrotic or apoptotic cells respectively [14]. In addition, activation-related receptors e.g. MHC and CD80/86 have been reported to be upregulated in the presence of necrotic cells [15]. We intend to extend and confirm these studies using proteomics, which will allow the pan-identification of novel, key proteins within the APC which are influenced by the presence of dying cells or 'danger signals'.

Hence, the aims of the immunological investigation can be summarised as
- To identify and investigate key APC-derived signals in response to co-culture with necrotic or apoptotic cells.
- To undertake functional analysis of the identified key signals in affecting the activation state of immune cells.
- To manipulate the co-culture system and derived signals upon results from the AIS / IDS studies.

### 2.3 Intrusion Detection Systems – Current State of the Art

An important and recent research issue for IDS is how to find true intrusion alerts from thousands alerts generated [19]. Existing IDS employ various types of sensors that monitor low-level system events. Those sensors report anomalies of network traffic patterns, unusual terminations of UNIX processes, memory usages, the attempts to access unauthorised files, etc. [24]. Although these reports are useful signals of real intrusions, they are often mixed with false alerts and their unmanageable volume forces a security officer to ignore most alerts [18]. Moreover, the low level of alerts makes it hard for a security officer to identify advancing intrusions that usually consist of different stages of attack sequences. For instance, hackers often use a number of preparatory stages (raising low-level alerts) before actual hacking [18]. Hence, the correlations between intrusion alerts from different attack stages provide more convincing attack scenarios than detecting an intrusion scenario based on low-level alerts from individual stages.

To correlate IDS alerts for detection of an intrusion scenario, recent studies have employed two different approaches: a probabilistic approach ([8], [36], [38]) and an expert system approach ([6], [7], [10], [33], [34]). The probabilistic approach represents known intrusion scenarios as Bayesian networks. The nodes of Bayesian networks are IDS alerts and the posterior likelihood between nodes is updated as new alerts are collected. The updated likelihood can lead to conclusions about a specific intrusion scenario occurring or not. The expert system approach initially builds possible intrusion scenarios by identifying low-level alerts. These alerts consist of prerequisites and consequences, and they are represented as hypergraphs ([33], [34]) or specification language forms ([6], [10], [18]). Known intrusion scenarios are detected by observing the low-level alerts at each stage. These approaches have the following problems [7]:

- Handling unobserved low-level alerts that comprise an intrusion scenario.
- Handling optional prerequisite actions and intrusion scenario variations.

The common trait of these problems is that the IDS can fail to detect an intrusion if an incomplete set of alerts comprising an intrusion scenario is reported. In handling this problem, the probabilistic approach is somewhat more advantageous because in theory it allows the IDS to correlate missing or mutated alerts. However, the similarities alone can fail to identify a causal relationship between prerequisite actions and actual attacks if pairs of prerequisite actions and actual attacks do not appear frequently enough to be reported.

Attackers often do not repeat the same actions in order to disguise their attempts. Thus, the current probabilistic approach fails to detect intrusions that do not show strong similarities between alert features but have causal relationships leading to final attacks ([8], [36], [38]).

## 3. A DT-Inspired Approach to Intrusion Detection

We propose AIS based on DT ideas that can handle the above IDS alert correlation problems. As outline previously, the DT explains the immune response of the human body by the interaction between APCs and various signals. The immune response of each APC is determined by the generation of danger signals through cellular stress or death. In particular, the balance and correlation between different signals depending on different causes appears to be critical to the immunological outcome. Proposed wet experiments of this project focus on understanding how the APCs react to the balance of different types of signals, and how this reaction leads to an overall immune response. Similarly, our IDS investigation will centre on understanding how intrusion scenarios would be detected by reacting to the balance of various types of alerts. In the HIS, APCs activate according to the balance of apoptotic and necrotic cells and this activation leads to protective immune responses. Similarly, the sensors in IDS report various low-level alerts and the correlation of these alerts will lead to the construction of an intrusion scenario.

### 3.1  Apoptotic versus Necrotic Alerts

We believe that various IDS alerts can be categorised into two groups: apoptotic type of alerts and necrotic type of alerts. Apoptotic alerts correspond to 'normal' cell death – hence, low-level alerts that could result from legitimate actions but could also be the prerequisites for an attack. Necrotic (unregulated cell death) alerts on the other hand relate to actual damage caused by a successful attack. An intrusion scenario consists of several actions, divided into prerequisite stages and actual attack stages [7]. For instance, in the case of Distributed Denial of Service (DDOS) intrusions, intruders initially look for vulnerable *Sadmind* services by executing the *Ping Sadmind* process [33]. This would be an apoptotic alert, relating to a prerequisite action. Just as apoptosis is vital in regulating the human immune response, apoptotic types of alerts are vital in detecting an intrusion scenario (since it indicates the prerequisite actions within an actual intrusion scenario) Necrotic alerts, or actual attack alerts are raised when the IDS observes system damage caused by the DDOS. Just as necrosis involves the unregulated cell death, necrotic types of alerts would be those generated from the unexpected system outcomes.

In our opinion, a better understanding how the APCs react to the balance of apoptotic and necrotic cells would help us to propose a new approach to correlate apoptotic and necrotic type of alerts generated from sensors. If the DT can explain the key proteins leading to necrotic and apoptotic signals, DT-based AIS would also be able to identify key *types* of apoptotic and necrotic alerts revealing the degree of alert correlation. In this way, DT-based AIS will correlate key types of alerts rather than specific alerts, and this will allow the AIS to correlate missing or mutated alerts as long as the key types of alerts are reported. For instance, in the DDOS example, an intruder can directly attack without executing *Ping* but executing the similar process *traceroute* instead [7]. In this case, our DT-based AIS should be able to link *traceroute* to DDOS attack damage since any type of scanning process is understood as an apoptotic type of alert for DDOS attacks.

### 3.2 Strength of Reactions

Additionally, if the DT can quantify the degree of the immune response, DT-based AIS would be able to quantify the degree of overall alert detection strictness. For instance, false positive alerts of IDS are often caused by inappropriately setting of intrusion signatures or anomaly thresholds. Debar and Wespi [9] use a clustering algorithm to group a large number of alerts and manually extract a generalised alarm reflecting each alarm cluster. By doing so, they identify the root cause of each alarm cluster and discriminate false positive alert clusters from true positive alert clusters. The root cause is the most basic cause that can reasonably be identified and fixed [9]. According to the identified root causes, new intrusion signatures or anomaly thresholds are redefined by removing those causing root causes.

However, their work has not reported further impacts after intrusion signatures and anomaly thresholds are reset. Simple removal of intrusion signatures that cause the root causes might degrade true positive detection rate instead. Furthermore, continuous changes of network and system environments require constant updates of intrusion signatures and anomaly thresholds. Thus, it is important for IDS how to react to false positive alerts and true positive alerts dynamically. The key feature of the DT-based AIS would provide a possible solution for this issue. The DT-based AIS would adopt a similar way to two types of immune cell death signals affecting the activation of nearby APCs. Currently observed balances between two types of alerts would affect the IDS sensors' activation status by resetting intrusion signatures or anomaly thresholds. Then, these new settings will result in new balances between the two types of alerts. If the DT can explain that this kind of cascading reaction stabilises in a way so that the overall immune responses can converge to an ideal status at given time, the DT-based AIS would also be able to follow a similar mechanism to identify the most suitable intrusion signature and anomaly thresholds setting at given time.

### 3.3   Danger Zones

Furthermore, our study aims to investigate how the danger alerts reported from a sensor can be transmitted to other sensors in order to detect on-going intrusions. Once a sensor has generated the danger signals or alerts, the AIS can quantify the degree of alert correlations indicating the strength of possible intrusion scenarios. If the AIS has strong indications of possible intrusion scenarios, it can activate other sensors that are spatially, temporally or logically 'near' the original sensor emitting the danger signal. This process is similar to the activated APCs sending its immune response providing a self-nonself independent grounding. For instance, when the danger signal reports the strong possibility of a web server compromise, this signal can be sent to other web servers in the same network domain.

## 4.   Summary and Conclusions

Our aim is to challenge the classical self-nonself viewpoint in AIS based IDS, and replace it by ideas from the DT. Existing systems using certain aspects of the HIS have been successful on small problems and have shown the same benefits as their natural counterparts: error tolerance, adaptation and self-monitoring. The DT is a new theory amongst immunologists stating that the natural immune system does not rely on self-nonself discrimination but identifies 'danger'. This is currently hotly debated by immunologists and far from widely accepted and has never before been applied to the IDS arena. It is our opinion that this theory is the key that will unlock the true potential of AIS by allowing us to build commercially viable systems that can scale up to real world problem sizes.

We intend to use the correlation of signals based on the DT. We believe the success of our system to be independent of the eventual acceptance or rejection of the DT by immunologist as the proposed AIS would achieve this by identifying key *types* of apoptotic and necrotic alerts and understanding the balance between these two types of alerts. In addition, the proposed AIS is extended by employing the APC activation mechanism explained by the DT. This mechanism has the advantage of detecting rapidly spreading viruses or scanning intrusions at an early stage.